\documentclass[sigconf, authorversion, nonacm]{acmart}

\AtBeginDocument{%
  }

\begin{document}

\title{Provenance Tracking in Large-Scale Machine Learning Systems}

\author{Gabriele Padovani}
\affiliation{%
  \institution{University of Trento}
  \city{Trento}
  \country{Italy}
}
\email{gabriele.padovani@unitn.it}

\author{Valentine Anantharaj}
\affiliation{%
  \institution{Oak Ridge National Laboratory}
  \city{Tennessee}
  \country{USA}}
\email{anantharajvg@ornl.gov}

\author{Sandro Fiore}
\affiliation{%
  \institution{University of Trento}
  \city{Trento}
  \country{Italy}
}
\email{sandro.fiore@unitn.it}

%
%
%
%

\renewcommand{\shortauthors}{Padovani et al.}

\begin{abstract}
As the demand for large-scale AI models continues to grow, the optimization of their training to balance computational efficiency, execution time, accuracy and energy consumption represents a critical multidimensional challenge.
Achieving this balance requires not only innovative algorithmic techniques and hardware architectures but also comprehensive tools for monitoring, analyzing, and understanding the underlying processes involved in model training and deployment. Provenance data—information about the origins, context, and transformations of data and processes—has become a key component in this pursuit. By leveraging provenance, researchers and engineers can gain insights into resource usage patterns, identify inefficiencies, and ensure reproducibility and accountability in AI development workflows.
For this reason, the question of how distributed resources can be optimally utilized to scale large AI models in an energy-efficient manner is a fundamental one.
To support this effort, we introduce the yProv4ML library, a tool designed to collect provenance data in JSON format, compliant with the W3C PROV and ProvML standards. yProv4ML focuses on flexibility and extensibility, and enables users to integrate additional data collection tools via plugins.
The library is fully integrated with the yProv framework, allowing for higher level pairing in tasks run also through workflow management systems. 
\end{abstract}





\maketitle

\section{Introduction}

In recent years Machine Learning (ML) has emerged as a transformative technology across a number of domains, such as, notably, climate change and energy efficiency \cite{eyring2024pushing}\cite{bracco2024machine}\cite{rolnick2022tackling}. 
However, as ML systems become more sophisticated and are deployed to address environmental challenges, there is an increasing need for transparency and interpretability too. In this context, provenance — \textit{the record of the origins, history, and transformations of data, models, and decisions}— \cite{cheney2009provenance} plays a crucial role. 
Provenance in ML should not be merely considered as a record-keeping practice; rather, it could also be instrumental in offering deeper insights into model behavior, data integrity, and the lifecycle of decision-making processes. 
From an AI model perspective, provenance could support a multi-dimensional approach, thus relying on an extensible set of orthogonal metrics of interest such as, among others, accuracy, computational time, and energy consumption. 

In such a context, the provenance of data and models could enable scientists and policymakers to interpret, analyse and compare the outputs of several ML systems \cite{liang2022advances}, knowing the methods and assumptions underpinning the results. Provenance could help in tracking the lineage of environmental data, model updates, and system parameters, ensuring that each stage of an ML-driven analysis can be validated, reproduced, and improved, thus increasing the level of trust in the overall AI experiment at large.

It has to be said, however, that the role of provenance in current ML implementation is still far from central. This issue stems from the fact that many ML systems prioritize performance optimization and scalability over interpretability, often leading to a trade-off where provenance tracking is overlooked to achieve faster model development and deployment \cite{jaigirdar2020information}. Additionally, the integration of provenance systems requires significant effort in terms of infrastructure and data management, and the recording and storing of such metadata further hits the efficiency. 

Lastly, the value of provenance is frequently not immediately apparent in commercial applications where short-term objectives may take precedence over long-term accountability and reproducibility \cite{kapoor2022leakage}. Consequently, despite its potential to enhance transparency, trust, and reliability, provenance remains an underutilized aspect of current machine learning practices.

The reasoning behind this under-utilization of this tool can be explained from many different fronts. Firstly, the shear amount of provenance data that can and needs to be collected to guarantee explainability is often performance impeding and extremely difficult for users to visualize. Secondly, the amount of additional work required by users to implement or integrate provenance tracking mechanisms is not trivial \cite{suriarachchi2015komadu}. Lastly, the benefits of tracking this large additional amount of data are often not immediate and may not be obvious to developers.

\begin{figure*}
    \centering
    \includegraphics[width=0.65\linewidth]{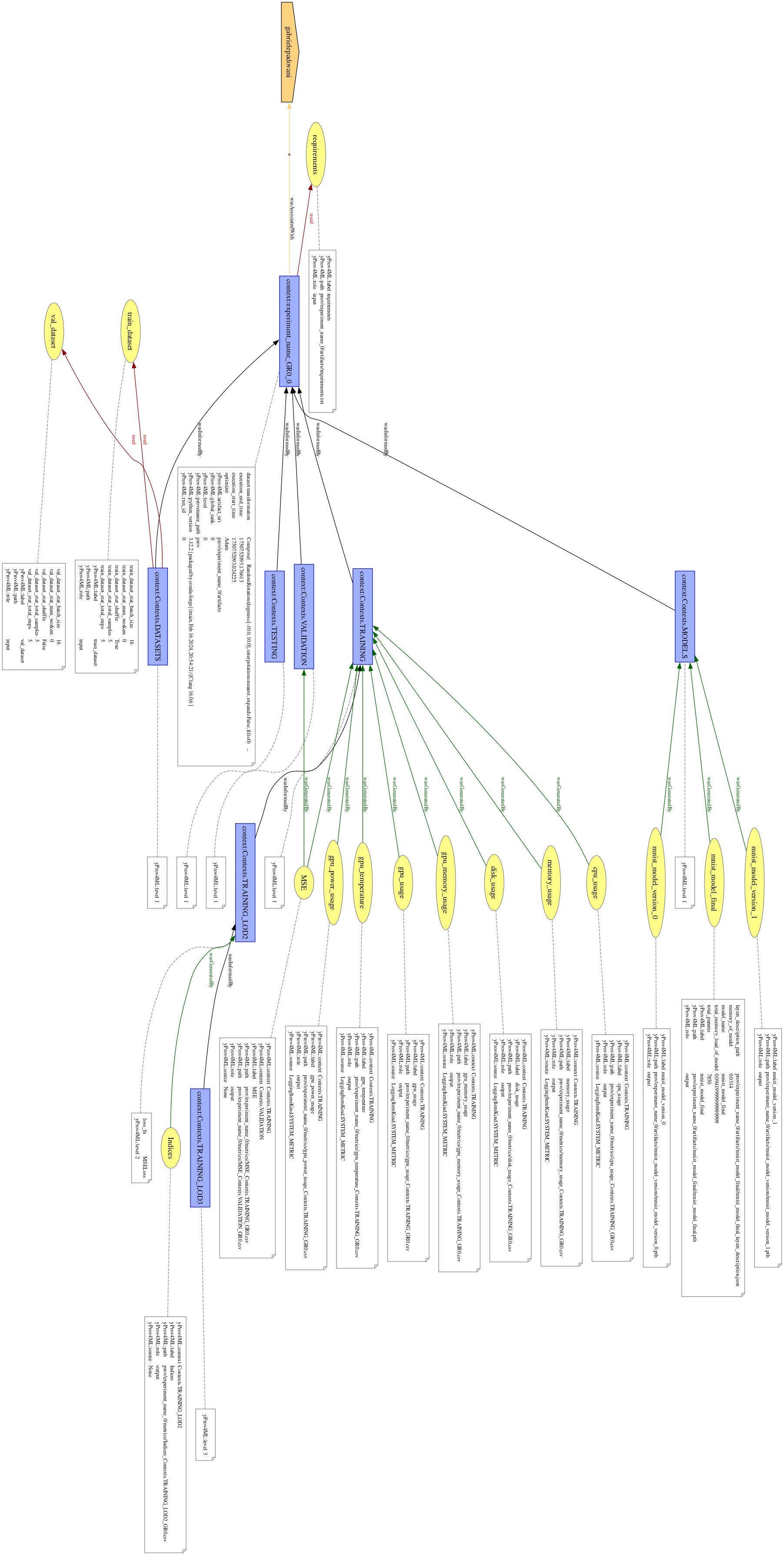}
    \caption{Example of provenance file created using the latest version of yProv4ML, it showcases the use of multiple contexts, and the creation of artifacts both as inputs (relationship "used") and outputs (relationship "wasGeneratedBy"). }
    \label{fig:prov-example}
\end{figure*}

For this reason, the yProv4ML\footnote{\url{https://github.com/HPCI-Lab/yProvML}} library was created, and aims at becoming a development tool for machine learning tasks, providing a set of logging utilities similar to MLFlow \cite{zaharia2018accelerating} and provides a recognizable interface for storing provenance data.
It additionally provides functionality for creating provenance files in PROV-JSON \cite{huynh2013prov} format, as well as provenance graphs, such as the one in Figure~\ref{fig:prov-example}. 

The aim of this work is to show the functionality of yProv4ML in conjuction with the surrounding yProv framework, which encompasses a set of provenance \textit{producers} \cite{sacco2024enabling}, such as the aforementioned library, and a set of \textit{consumers}, such as the provenance handle system and the yProvExplorer, both of which will be described in the following sections. In addition, a set of use cases are have already been shown in \cite{padovani2024software}, and an additional one will described in the later section of this paper. 
Since many of use cases of yProv4ML have been showcased, an emphasis will be put on the new functionalities, introduced in the latest library version\footnote{\url{https://github.com/HPCI-Lab/yProvML/tree/development}}. 

This document is structured in the following sections: \textit{Introduction}, where the main topic and issues presented are defined; \textit{Related Work}, where the context of this research is explored, along with the prominent works and the libraries already estabilished in this field; \textit{Applications and Functionality} shows a set of possible use cases and where provenance collection can be useful during the development process of machine learning systems; \textit{Use cases}, which explores a real research scaling study of a foundation model, developed in collaboration with the Oak Ridge National Laboratory; and finally \textit{Conclusions}. 

\section{Related Work}

The tracking of provenance has become an increasingly essential component of complex data systems, enabling transparency, reproducibility, and accountability across diverse workflows. Various standards and models have been proposed to address these needs, with the World Wide Web Consortium (W3C\footnote{https://www.w3.org}) leading efforts through the PROV family of standards \cite{missier2013w3c} to support the gathering and exchange of provenance information.
The W3C PROV family offers a framework that supports consistent representation of data origin and lineage. This framework comprises several core components: PROV-DM \cite{belhajjame2013prov}, which serves as a relational data model to encapsulate key elements of provenance; PROV-O \footnote{https://www.w3.org/TR/prov-o/}, an OWL ontology that enables the conceptual modeling of provenance data; and PROV-N \cite{moreau2013prov}, a syntax designed to be both relational and human-readable, enhancing accessibility for users.

The principal goal of W3C PROV is to foster interoperability, making it possible for different provenance-producing systems to exchange structured information seamlessly. This has led to its adoption across various domains, with several systems extending W3C PROV for specific use cases. For instance, PROV-ML \cite{souza2019provenance}\cite{souza2019efficient} applies W3C PROV standards within the scientific machine learning lifecycle, incorporating elements from both W3C PROV and the ML Schema (MLS) \cite{publio2018ml}. It is based on the creation of a taxonomy comprising three predetermined phases: data curation, data preparation, and learning. However, this approach has limitations. It restricts end-users to a set of specific phases and does not allow for full flexibility in the ML pipeline. 

Another example is ProvLake \cite{souza2022workflow}, a system that captures provenance data across diverse workflows, adapted to support PROV-ML and handle complex queries related to the ML lifecycle. Additionally, MLflow2PROV \cite{schlegel2023mlflow2prov} facilitates the extraction of W3C PROV-compliant provenance graphs from MLflow-managed experiments, combining these with data from Git repositories to create comprehensive provenance representations for machine learning projects~\cite{10.1162/dint_a_00119}.

While W3C PROV supports a wide range of applications, the moltitude of applications in which provenance can be leveraged, makes interoperability fundamental. For this reason, the PROV-JSON~\cite{niu2015interoperability} standard has been developed, which makes use of the human-readable and open format JSON. 

In addition to these W3C PROV-based examples, several other papers present cutting-edge applications of provenance in Machine Learning. 
As for the context of deep learning \cite{pina2020provenance}, uses provenance to support the analysis of hyperparameters in neural networks, facilitating the optimization of the learning process. A similar approach is brought forward in \cite{khan2019data}, where a provenance-based system for distributed machine learning environments based on StellarGraph\footnote{https://www.stellargraph.io/} is proposed. 
As for a more fine-grained approach, the LIMA framework \cite{phani2021lima} can be used for tracking and reusing provenance data in machine learning systems. LIMA focuses on reducing computational redundancy by tracking data at the level of individual operations and integrates with techniques for parallelism and merging operators. For an explainability-oriented approach, on the other hand, \cite{waltersdorfer2024auditmai} explores continuous auditability infrastructure in the context of federated agencies integrating machine learning systems. 

Shifting the focus on a more workflow-oriented approach, Workflow Run RO-Crate \cite{leo2024recording} is an extension of the RO-Crate model \cite{soiland2022packaging} to record the provenance of workflow executions. Workflow Run RO-Crate, based on W3C PROV, aims to improve interoperability between different workflow management systems. 

On the other hand, \cite{souza2023towards} examines the function of workflow provenance in facilitating the advancement of reliable, and energy-efficient AI, from edge devices to cloud and high-performance computing (HPC). It focuses on handling heterogeneous data and capturing human interactions, while also managing energy consumption, both in the context of ML tasks and in larger workflows \cite{souza2024workflow}. 

Despite its importance, current ML tracking solutions exhibit several significant gaps. The former challenge is posed by scalability, as ML experiments can grow in complexity and scale very rapidly, and existing tracking systems may struggle to handle the increased volume and granularity of data. It is therefore critical to manage ML experiments in a lightweight manner in order to avoid performance bottlenecks and enable large-scale processes.
Approaches such as Weights and Biases \cite{wandb} resolve this issue by streaming directly all logged information to the cloud, which allows for a more lightweight footprint on the local architecture. 
The second significant challenge arises from the lack of interoperability. While numerous machine learning frameworks have been widely adopted, the vast majority generate provenance information in proprietary formats, which presents a challenge when having to share data or mix tools. This lack of standardization makes it difficult to aggregate, compare, and share provenance data across platforms, limiting the usefulness of this type of information in collaborative and cross-disciplinary research.

To address this issue, a potential solution could be the development of third-party tools designed to track data provenance across disparate environments and integrate the metadata from these systems into a unified representation. 
To this end, the yProv \cite{fiore2023graph} framework tries to address provenance challenges in a more holistic manner, from management, tracking, analysis and exploration through a seamless, consistent and interoperable software ecosystem. 
yProv is an provenance management service that conforms to the W3C PROV family of standards, leveraging a graph data model and exposing a RESTful API \cite{fiore2023graph} for a range of use cases \cite{padovani2024software}.
From a software architecture perspective, the yProv service consists of three main components: the yProv web service front-end; a graph database engine back-end based on Neo4J; and the yProv command line interface (CLI), which provides a set of commands for invoking the RESTful APIs.
While the yProv service and the yProv Explorer\footnote{\url{https://explorer.yprov.disi.unitn.it/?file=http\%3A\%2F\%2Fyprov.disi.unitn.it\%3A3000\%2Fapi\%2Fv0\%2Fdocuments\%2Fsoftwarex\#model_name}} can be considered provenance \textit{consumers},
meaning that they accept a PROV-JSON file and elaborate its information, the yProv4ML library, 
as well as its workflow counterpart yProv4WFs, are provenance producers. For the latter class,
we focused on keeping the PROV-JSON files created as generalized as possible, meaning avoiding 
domain-oriented tags, activities and entities. 

\section{Applications and Functionality}

In the context of provenance collection and utilization, four initial use cases are identified, 
which will be expanded upon in the following sections.
In particular, these will focus on expanding the development helper functionalities of yProv4ML,
and show what are the benefits of collecting, with minimal overhead, this type of information.

\subsection{Development tracking}

The creation of datasets for machine learning frequently entails the utilization of numerous ad hoc and repeated transformations, where information is modified in a manner that renders it comprehensible to these programs. Challenges emerge when revisiting these steps to reproduce and trace the origin of the process. In such instances, the transformations are often lost, resulting in slight divergences from the preceding design process. 
It is of critical importance to document the complete design workflow in order to ensure a comprehensive reproduction of experiments, as well as to prevent the generation of unreplicable results. 

Similarly, the definition of a development environment follows a similar process, with a multitude of commands which are executed in an arbitrary manner, and may be lost or forgotten by the author. 

For this reason, the plan with yProv4ML is to collect first of all the output of the training script, to fully understand what was the outcome of the program's execution, and secondly to compress the full list of executed console commands, along with the textual output of each one, obtaining a "development graph" of all changes which the programming environment was subjected to. 

In the same set of experiments, additionally, the possibility of tracking git differences could be enabled, increasing the level of detail of the tracked information and avoiding ad-hoc modifications. 
This could also enable a one-to-one memorization of each modification, along with the results obtained for the specific version of the script, allowing the user to investigate which version of the project worked better. 
A developer would be able to see all modifications and revert changes to the extent of the exact run, as long as the data was saved. 
This also enhances reproducibility and explainability, as the user would be able to easily roll back to a specific moment in time and understand what caused the change. 

\subsection{Trade-offs oriented training}

Many computing centers allocate a fixed amount of node hours for projects, setting a cap on the computational resources available. Consequently, if a team of researchers wanted to identify the optimal model achievable, they would be required to conduct as few runs as possible when developing the architecture and running preliminary experiments. With a knowledge base of previous runs available and metadata easily searchable, the team could identify similar processes and have a rough estimate of parameters to base their tests on. 

This issue is made even more critical when working with foundation models or other large deep learning models. The addition of modules is necessary for the scaling of the number of parameters of a model, but there is currently no way of estimating how this change could impact performance. With enough fine-grained provenance information, the forecasting of the impact of a specific change could be discerned, assuming to have a baseline of the training outcome, and a list of the changes conducted to the code base. 

In addition, an online provenance tracking process could give real-time guidelines in how to proceed during the training process, understanding when to stop. This would result in a more optimized use of compute hours, as the process could be stopped when a specific threshold of energy, compute, or performance is achieved, removing unnecessary iterations. 

\subsection{Scaling studies performance estimation without training}

As anticipated in the previous section, in the context of large-scale deep learning projects, especially recently with the increased popularity of foundation models, it is often necessary to perform scaling studies to understand how the model behaves when increasing its parameters, the dataset size, the amount of FLOPs expended, or the number of computing devices \cite{yin2023evaluation}. 
Especially when dealing with architectures such as foundation models on a distributed scale, it can be difficult to achieve a stable and convergent behaviour, and this often results in a large number of similar experiments with a set of increasingly larger parameters. 

Having access to a dataset that contains fine-grained information about similar applications could help to understand how the architecture would behave when increasing a particular parameter, without having to train the model from scratch each time \cite{kaplan2020scalinglawsneurallanguage}\cite{hoffmann2022trainingcomputeoptimallargelanguage}\cite{muennighoff2023scalingdataconstrainedlanguagemodels}. 

In addition, this issue becomes even more problematic when considering longer machine learning training pipelines, such as those where a dataset is preprocessed prior to model fitting. In this case, identifying possible beneficial transformations without having to re-run the entire pipeline guarantees not only a faster development process, but also a more sustainable one, that consumes less energy and resources. 

On this topic, two approaches to solve this could be tackled. The former utilizes an analytical approach to determine an estimate of the performance when scaling one of the three aforementioned factors. This could give a precise estimate of both compute necessary and the configurations of architecture adoptable to maximize performance. 
Another approach, on the other hand, would revolve around the use of historical data from previous, but similar, experiments. A ML-based forecasting approach could give, this way, a more precise estimate of any of the pivotal factors. To give more context, the model in question would understand what is the contribution of each component present in the ML pipeline, in the context of only similar experiments, and would be able to give a prediction with a single inference step, eliminating the trial and error phase of adding and removing modules or of tweaking parameters. 

Integrating yProv4ML would allow precise tracking and comparison of runs, meaning that a user could go back and identify the most optimal combination of parameters, or even infer the best combination, either from current runs, or from other developers' runs. 

\subsection{Hyperparameter tuning}

While the former section explained a possible improvement to the development pipeline, by allowing users to pair project differences with results, a better approach could revolve around the grouping of the results of a high number of experiments. This way, users will be able to identify targets that are similar to their own and deduce the optimal hyperparameter values for their particular application. 

A further issue with hyperparameter tuning arises when numerous attempts at training the optimal model are made through trial and error. When the same process is repeated several times, a great deal of computing resources are expended unnecessarily. This is particularly problematic when dealing with machine learning models that are large in size, as the aforementioned paradigm quickly becomes unsustainable, especially when the models in question consist of billions of parameters. 
Adding to this, preprocessing pipelines in ML tend to also run for a long time and require parameter tuning of their own. 

\section{Methodology}

In order to tackle the aforementioned issues, a provenance collection library, yProv4ML, has been developed. 
As mentioned before, this module is intended to be a part of the yProv suite of tools, which allows to collect, upload and explore provenance files used in a variety of tasks. 

\begin{figure}
    \includegraphics[width=\linewidth]{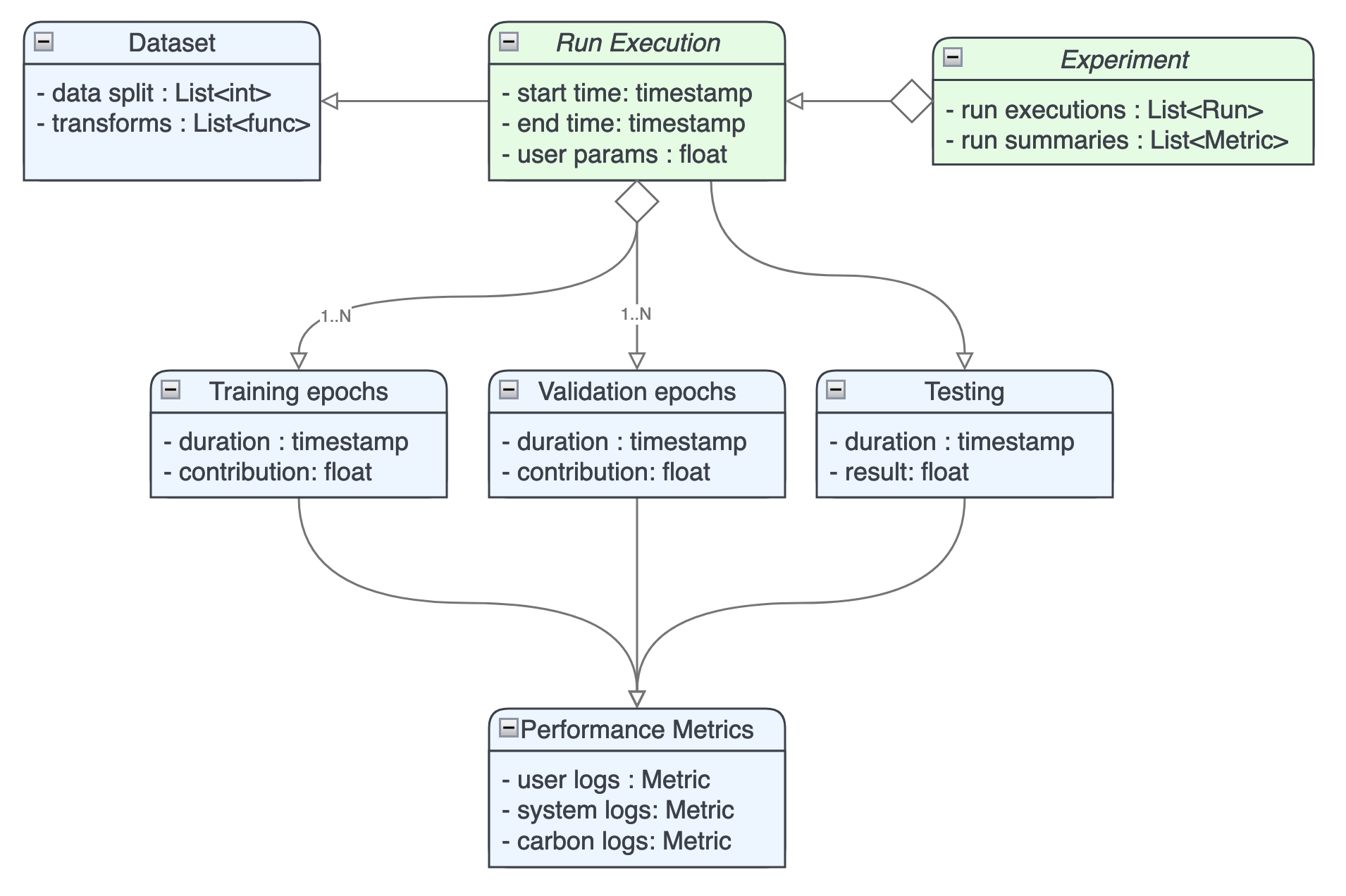}
    \caption{Data Model used as foundation for yProv4ML, blue blocks identify contexts and green ones are related to experiments and individual runs. }
    \label{fig:prov4ml-datamodel}
\end{figure}

Figure \ref{fig:prov4ml-datamodel} illustrates the yProv4ML data model. The core entity in this model is an \textit{Experiment}, which includes different \textit{Run Execution} instances. This structure allows for multiple runs under a single experiment, each potentially configured with different parameters, making it possible to systematically compare results across multiple executions.
In each run, the process is divided into stages, called contexts, including training, validation, and testing, while others can be defined by the user. The training and validation stages, in this case, are organized into epochs, each of which captures specific details such as duration and contribution to the final model. 

The Prov4ML library exposes logging utilities similar to MLFlow, allowing for quick integration and researchers to focus on their work. It collects three main categories of information: artifacts, parameters, and metrics. 
The former identifies any file or output that may be used later in the next phases of the workflow. For machine learning processes, these typically include model versions, checkpoints, and source code material. 
Parameters, on the other hand, are one-time recorded values used during training. Some examples include learning rate, model size and width, and other hyperparameters. 
The last category contains information that is updated during the training process. These metrics include losses and program execution statistics such as energy efficiency, power consumption, and GPU usage. 

Once information about a single run is stored, it is also possible to compare the results of successive, related runs. This allows for a better understanding of the impact of hyperparameters and model configurations, while keeping track of any changes made to the entire script. 

The core idea behind this library, is to be able to log more lineage information compared to MLFlow or Weights and Biases, so to allow the user to track at a much deeper level the entirety of the ML process. In addition to the more fine-grained data resolution, yProv4ML works along side MLFlow, so to offer a standardized pipeline through which to log data, allowing the user to modify minimal portions of code. 
While the plan for this library is not to take over or to be a competitor to MLFlow, it is evident that provenance data could help significantly in the same tasks the aforementioned library, and the integration of a plugin to allow for integration between the two is already in the works. 
This process would allow for the collection of a large amount of standardized ML pipelines, which could then be used to conduct performance inference, scaling studies, and other analyses, all with minimal resource consumption and focusing on the contribution of each small component. 

Where by default the output of the library is a single provenance file, containing all parameters, metrics and links to stored artifacts, the newest version of yProv4ML allows for storing of time series and bulky data into additional provenance files, taking advantage of more advanced open file formats, such as NetCDF \cite{rew1993netcdf} and Zarr\cite{zarr}. This allows to keep the provenance file describing the process at a top level as small as possible, while also providing a suite of surrounding tools and data easily accessible to the user. 
Storing metrics in a more optimized file (such as Zarr), avoids large overhead when having to store numerical data in text format, when passing through JSON. 
Preliminary work on this idea show gains of more than 90\% on average. More of these results are shown in Table \ref{tab:zarr-prov}. 

\begin{table}
    \centering
    \caption{Provenance file size comparison in normal and compressed formats, the measurements include both the PROV-JSON and the additional metric files, which are separated in Zarr and NetCDF. }
    \begin{tabular}{|l|c|c|}
        \hline
        \textbf{File} & \textbf{Normal Size} & \textbf{Compressed Size} \\
        \hline
        Original\_file.json & 39.82 MB & 8.65 MB \\
        Converted\_to.zarr & 2.74 MB & 2.14 MB \\
        Converted\_to.nc & 2.35 MB & 2.30 MB \\
        \hline
    \end{tabular}
    \label{tab:zarr-prov}
\end{table}

In addition to the former changes, the input and outputs relationships in provenance files have been reworked. When defining metrics, artifacts and parameters, it is now possible to define whether the data logged is an input, otherwise defaulting to an output. This allows to identify parameters which are necessary for the execution, as well as the ones which will have to be reproduced by the end, in case some user decided to re-run a specific experiment, described through a provenance json file. 
While it is currently not yet achievable, with the help of a git commit and a set of scheduling commands, we aim to guarantee reproducibility out-of-the-box when using yProv4ML. With this change, reproducing an experiment by simply sharing a provJSON file would become trivial, and would be a step towards the creation of a trustworthy provenance infrastructure \cite{marchioro2025trustworthy}.

In addition to the previous changes, since the W3C Prov standard is not the only one used, yProv4ML will now allow to create a wrapper around the artifact directory using RO-Crates, which guarantees self-describing capability when having to share a single experiment.
A quick summary of the functionalities of each standard is presented in the Table~\ref{tab:rocrate}. While the former is a general-purpose framework designed to represent the provenance of digital and physical entities, the latter is a lightweight packaging format that bundles data, and in many use cases a more specific format might be preferred. 

\begin{table*}[h!]
\centering
\caption{Comparison between the W3C PROV standards and RO-Crate specific. The utilization of both is summarized in the last row of the table. }
\label{tab:rocrate}
\begin{tabular}{|l|l|l|}
\hline
\textbf{Feature} & \textbf{W3C PROV} & \textbf{RO-Crate} \\
\hline
Type & Provenance data model & Research object packaging format \\
\hline
Standardized By & W3C & Community-driven \\
\hline
Serialization & PROV-N, PROV-JSON, PROV-O (RDF) & JSON-LD \\
\hline
Focus & Provenance representation & Sharing and describing research artifacts \\
\hline
Packaging & No & Yes \\
\hline
Domain-Agnostic & Yes & Can be \\
\hline
Use of W3C PROV & Native & Optional (via PROV-O) \\
\hline
\textbf{Use in yProv4ML} & \textbf{Tracking of provenance} & \textbf{Packaging of artifacts} \\
\hline
\end{tabular}
\end{table*}

\section{Use case}

The aforementioned collaboration with the Oak Ridge National Laboratory team revolved around the creation of a foundation model (MODIS-FM) utilizing a Vision Transformer (ViT) architecture \cite{alexey2020image}, aimed at atmospheric remote sensing applications. The name MODIS-FM is derived from NASA’s Moderate Resolution Imaging Spectroradiometer (MODIS) \cite{platnick2016modis}, which provided the data for its training. The fitting process was executed on the Frontier supercomputer at the Oak Ridge Leadership Computing Facility, which consists of 9,402 compute nodes, each equipped with a 64-core AMD EPYC CPU and 8 AMD Instinct MI250X Graphics Compute Dies (GCDs), effectively functioning as a single GPU.

In order to train a large final model, several scaling studies were conducted to understand which configuration of parameters would be more adequate to be used for the training process, as well as how the architecture would behave when increasing its size, the amount of data, and the amount of computing devices. 

Two model baselines were taken into consideration for the studies, the former being a Swin Transformer V2 \cite{liu2022swin}, while the latter a Masked Autoencoder with a vision transformer (ViT) baseline \cite{he2022masked}.
The former models was chosen as it offers great performance for the amount of computation, while the latter features masked training, which is considered very helpful in the context of climate forecasting.  

This study utilizes 23 years of MODIS 1km L1B radiance data obtained from NASA's Aqua and Terra satellites, resulting in around 800.000 128x128 patches, each with 6 channels consisting of a specific atmospheric variable. 
Four model configurations were identified for the scaling studies, with 100 million, 200 million, 600 million, and 1.4 billion parameters respectively. For each of these models, experiments were run for five different configurations of compute devices: 8, 16, 32, 64, and 128 GPUs respectively, following the Distributed Data Parallel (DDP) \cite{li2020pytorch} training approach. In the pre-training stage, the model undergoes self-supervised training, where the transformer is tasked with reconstructing the input patch. On the other hand, in the fine-tuning stage, all layers except for the final prediction head are kept frozen, and the model is trained using labeled data.

\begin{figure*}
    \centering
    \includegraphics[width=0.48\linewidth]{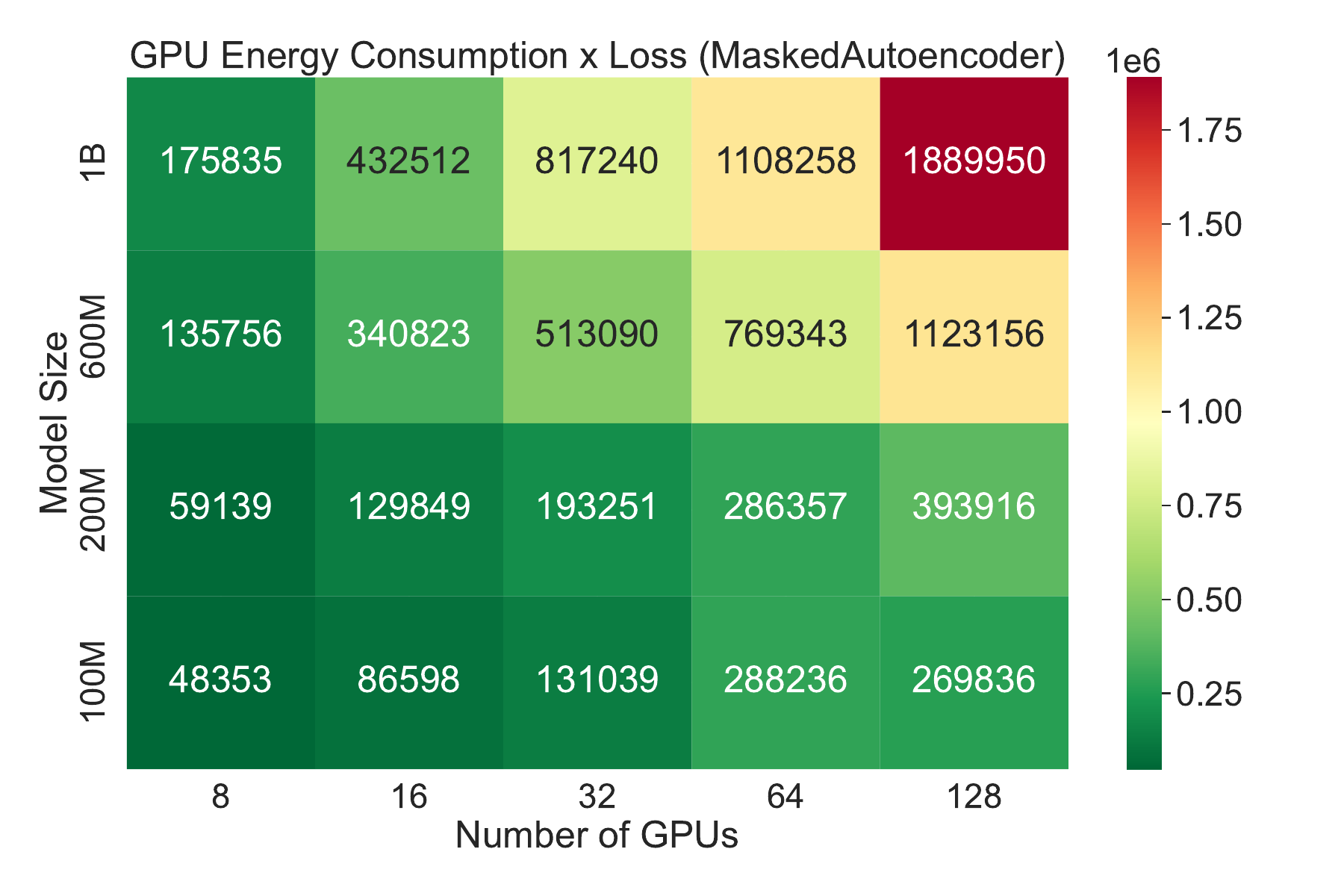}
    \includegraphics[width=0.48\linewidth]{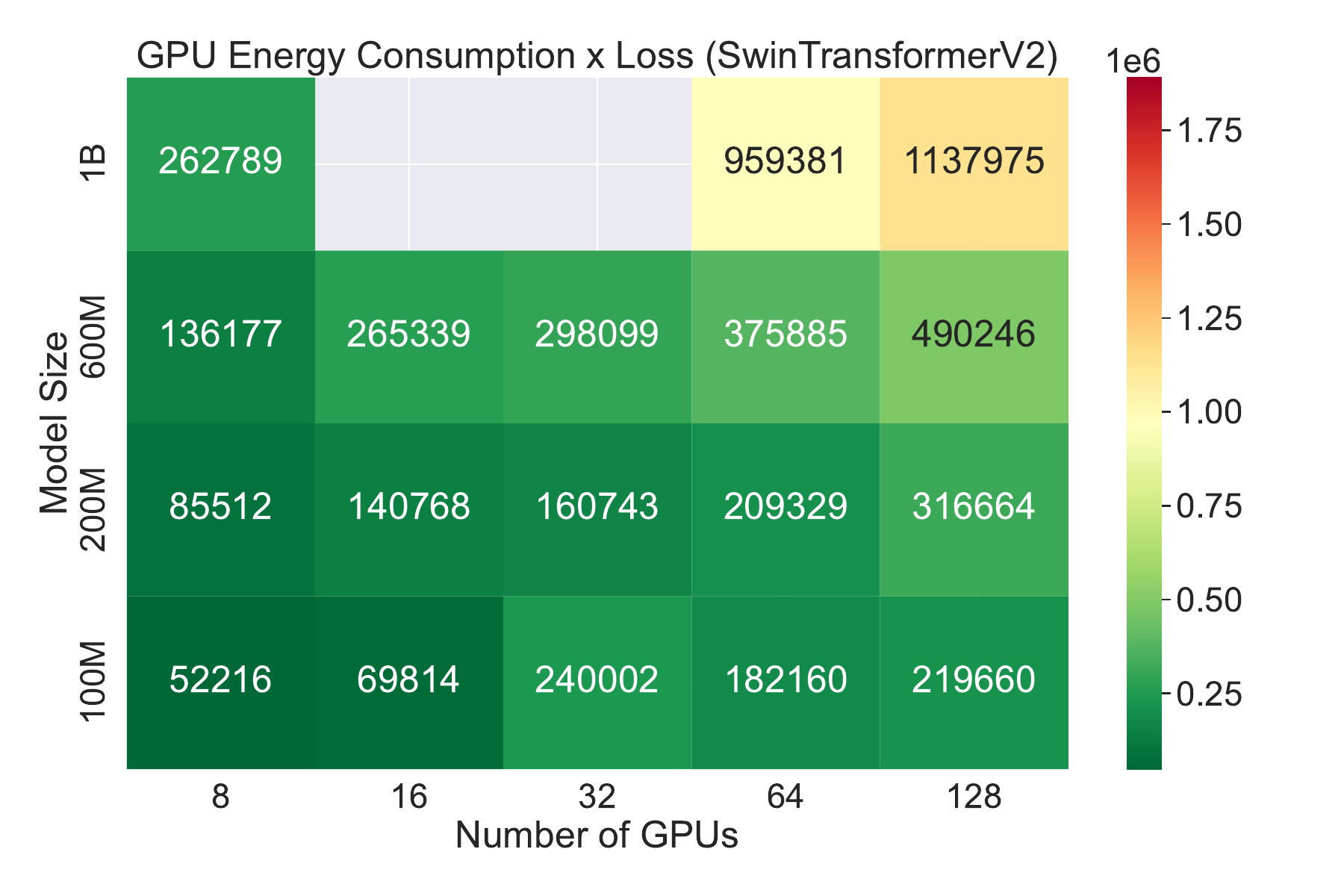}

    \caption{Energy and performance trade-off, calculated as the loss times the total energy consumption, for MAE (top) and SwinT (bottom). Empty cells indicate experiments which ran for longer than the 2 hours walltime. }
    \label{fig:energy-loss}
\end{figure*}

Results in Figure \ref{fig:energy-loss} show that the best trade-off between energy consumption and performance is obtained. Although it is clear that a smaller model and smaller compute are beneficial when the dataset is contained, when scaling up the samples used it becomes unreasonable to stick with less compute devices. 
When comparing the two models under this aspect, the newer SwinT-V2 architecture is performing much better at scale, while MAE presents a steeper trade-off curve.

\section{Conclusions and Future work}

Optimizing energy efficiency while improving reproducibility and scalability is a topic that with time is increasingly gaining following. By leveraging provenance data, the yProv4ML library will enable researchers to track and analyze the entire lifecycle of ML models at low levels of granularity, facilitating efficient resource utilization. We believe that having a dedicated provenance collection library, designed to capture ad-hoc information from the training phase is a fundamental step in guaranteeing more trustworthy results from the community. This supports in-depth scaling studies, as well as provides ML researchers with valuable tools for sustainable and reproducible research practices.

Future work on this library will target the development aspect of machine learning, by tracking all experiment runs in a single provenance files, to enable easier comparison with each individual execution. In addition, the reproducibility aspect will be expanded, by enabling researchers to reconstruct use cases using a single PROV-JSON file.  

As ML models become increasingly complex, helping users balance performance with energy considerations becomes a task of even greater importance. The project's collaborations with ORNL demonstrate its applicability to diverse domains, from atmospheric science to cybersecurity. Ultimately, this work has the potential to change the way distributed ML models are trained, promoting an approach that is more environmentally sustainable and reproducible, limiting the number of runs and experiments.

\begin{acks}

\textit{This work was partially funded under the National Recovery and Resilience Plan (NRRP), Mission 4 Component 2 Investment 1.4 - Call for tender No. 1031 of 17/06/2022 of Italian Ministry for University and Research funded by the European Union – NextGenerationEU (proj. nr. CN\_00000013) and the EU InterTwin project (Grant Agreement 101058386).}

\textit{Moreover this research used resources of the Oak Ridge Leadership Computing Facility at the Oak Ridge National Laboratory, which is supported by the Office of Science of the U.S. Department of Energy under Contract No. DE-AC05-00OR22725.}

\end{acks}

\bibliographystyle{unsrt}
\bibliography{bibliography}

\end{document}